\documentclass[letterpaper]{article} 
\usepackage{aaai2026}  
\usepackage{times}  
\usepackage{helvet}  
\usepackage{courier}  
\usepackage[hyphens]{url}  
\usepackage{graphicx} 
\usepackage{natbib}  
\usepackage{caption} 
\usepackage{algorithm}
\usepackage{algorithmic}
\usepackage{xcolor}
\newcommand{\answerYes}[1]{\textcolor{blue}{#1}} 
\newcommand{\answerNo}[1]{\textcolor{teal}{#1}} 
\newcommand{\answerNA}[1]{\textcolor{gray}{#1}} 
 
\usepackage{newfloat}
\usepackage{listings}
\DeclareCaptionStyle{ruled}{labelfont=normalfont,labelsep=colon,strut=off} 
\floatstyle{ruled}
\newfloat{listing}{tb}{lst}{}
\floatname{listing}{Listing}
\title{ Mapping the Political Discourse in the Brazilian Chamber of Deputies:\\ A Multi-Faceted Computational Approach}

\author{
    Flávio Soriano\textsuperscript{\rm 1},
    Victoria F. Mello\textsuperscript{\rm 1},
    Pedro B. Rigueira\textsuperscript{\rm 1},
    Gisele L. Pappa\textsuperscript{\rm 1},
    Wagner Meira Jr.\textsuperscript{\rm 1},
    Ana Paula Couto da Silva\textsuperscript{\rm 1},
    Jussara M. Almeida\textsuperscript{\rm 1}
}
\affiliations{
    \textsuperscript{\rm 1}Computer Science Department, Universidade Federal de Minas Gerais, Belo Horizonte, Minas Gerais, Brazil\\

    flaviosoriano@dcc.ufmg.br, victoriaflores@dcc.ufmg.br, pedrobacelar.rigueira@dcc.ufmg.br, glpappa@dcc.ufmg.br, meira@dcc.ufmg.br, ana.coutosilva@dcc.ufmg.br, jussara@dcc.ufmg.br

}

\usepackage{booktabs}
\usepackage{appendix}

\usepackage{color}

\begin{document}

\maketitle

\begin{abstract}
Analyses of legislative behavior often rely on voting records, overlooking the rich semantic and rhetorical content of political speech. In this paper, we ask three complementary questions about parliamentary discourse: how things are said, what is being said, and who is speaking in discursively similar ways. To answer these questions, we introduce a scalable and generalizable computational framework that combines diachronic stylometric analysis, contextual topic modeling, and semantic clustering of deputies’ speeches. We apply this framework to a large-scale case study of the Brazilian Chamber of Deputies, using a corpus of over 450,000 speeches from 2003 to 2025. Our results show a long-term stylistic shift toward shorter and more direct speeches, a legislative agenda that reorients sharply in response to national crises, and a granular map of discursive alignments in which regional and gender identities often prove more salient than formal party affiliation. More broadly, this work offers a robust methodology for analyzing parliamentary discourse as a multidimensional phenomenon that complements traditional vote-based approaches.
\end{abstract}

\begin{links}
\link{Code}{https://github.com/flaviosoriano/multifaceted-political-analysis}
\end{links}

\section{Introduction}

Political speeches are a central medium through which representatives communicate agendas, negotiate positions, and influence public perception. Analyzing these speeches provides insight into both the institutional dynamics of legislatures and the broader political landscape. In Brazil, one of the largest democratic countries in the world, the Chamber of Deputies, the lower house of the National Congress, is the main institutional space for legislative debate and political representation. In this institution, elected representatives from across the federation articulate the interests of their constituents, formulate national-level policies, and oversee the Executive Branch. The primary outcome of this deliberative process is an extensive public record of parliamentary speeches. These records are not merely procedural transcripts; they constitute the formal materialization of political debate, allowing for the observation of how governmental priorities are defined, ideological positions are confronted, and the country's direction is discussed.

Analyzing this massive volume of political discourse is crucial for understanding fundamental democratic processes, including political polarization, agenda-setting, and the formation of partisan alliances. While traditional computational analyses of legislative bodies often rely on structured data like voting records to map political alignments \cite{poole1985spatial, Izumi2023}, such approaches overlook the rich semantic and rhetorical dimensions of the deliberative process itself. They reveal the outcomes of debates, but not the arguments, semantic frames, and discursive strategies that shape those outcomes. Understanding this deliberative dimension, namely how politicians use language to persuade, define issues, and position themselves, is essential for a deeper and more nuanced view of legislative politics. 

While a few initiatives, particularly in the U.S. Congress context, have made textual records of debates (e.g., the Congressional Record) available and spurred follow-up studies using NLP methods, the literature remains limited in its application of state-of-the-art techniques to political speeches. Existing contributions have primarily focused on data collection and curation \cite{judd2017, bochenek2025}, or have applied relatively narrow methodological approaches \cite{rodriguez2022word}  or have focused on specific topics (e.g., immigration) \cite{card2022}. As a result, they stop short of offering a comprehensive, discourse-centered analysis


This paper takes the next step toward a systematic analysis of political speeches by treating parliamentary discourse as a multidimensional phenomenon. More specifically, we ask three complementary questions: how is parliamentary discourse constructed, what themes organize it over time, and who speaks in discursively similar ways. We address these questions through a unified NLP pipeline applied to the Brazilian Chamber of Deputies. First, we conduct a diachronic stylometric analysis to capture how the structure and rhetorical form of speeches evolve over time. Second, we use contextual topic modeling to identify what is being said, that is, the thematic agenda that organizes parliamentary debate. Third, we leverage state-of-the-art document embeddings and clustering to analyze who speaks similarly in semantic space, revealing discursive alignments that may or may not coincide with formal party boundaries.

Applying this methodology to a comprehensive and publicly available corpus of over 450,000 speeches from 2003 to 2025, our work makes several key observations. First, our thematic analysis reveals a highly reactive legislative agenda, with dominant topics shifting abruptly in response to national crises, including the 2016 impeachment of president Dilma Rousseff\cite{nyt_dilma_impeachment}, and the COVID-19 pandemic. Second, by mapping the semantic space, we identify granular ideological clusters that transcend formal party lines and trace the diachronic trajectories of political parties, revealing patterns of convergence and divergence. Third, our stylometric analysis uncovers a long-term trend towards shorter and structurally simpler speeches, which is consistent with the broader mediatization of political communication. Together, these findings provide a data-driven account of the modern Brazilian political landscape, offering also a robust and scalable methodology for understanding the linguistic dynamics of an important democratic institution.

The three analytical components of our framework are necessary because they capture non-redundant dimensions of parliamentary discourse. Stylometric analysis reveals how speeches are constructed, allowing us to track shifts in rhetorical form and communicative style over time. Topic modeling reveals what is being said, identifying the substantive agenda that structures parliamentary debate and how it changes in response to political events. Semantic clustering reveals who speaks in discursively similar ways, mapping alignments among deputies that may coincide with, but also diverge from, formal party organization. Taken together, these analyses allow us to study parliamentary discourse as a multidimensional phenomenon of form, content, and alignment, rather than reducing it to any single textual property.

In sum, our main contributions are:
\begin{itemize}
\item We propose a multidimensional framework for analyzing parliamentary discourse through three complementary questions: how it is said, what is being said, and who speaks in discursively similar ways. This extends legislative studies beyond structured records to capture the form, content, and alignment of political speech.
\item We operationalize this framework through a unified NLP pipeline that combines diachronic stylometric analysis, contextual topic modeling, and semantic clustering, providing a scalable methodology for the large-scale study of legislative speech.
\item We apply this methodology to the speeches of Brazilian deputies over a 22-year period, offering a large-scale empirical characterization of parliamentary discourse in a major democratic country that remains comparatively underexplored in the literature.
\end{itemize}

\section{Related Work}
\label{sec:related_work}

The quantitative analysis of political positioning in legislative bodies has traditionally relied on roll-call votes to infer the ideological alignment of parliamentarians in a latent space \cite{poole1985spatial, PaperCarlos,  Izumi2023}. These foundational methods are powerful for mapping party cohesion and the main dimensions of political conflict. Yet, they are inherently limited as they only capture the final outcome of deliberation, remaining silent on the rich processes of justification, persuasion, and issue framing that shape political decisions.

The ``text-as-data'' movement emerged as a direct response to this gap, providing a computational toolkit to extract political positions directly from language \cite{Laver2003}.  Indeed,  a few prior studies have focused  on making the U.S. Congressional Record available for research, by offering crawling and parsing tools \cite{judd2017} or curated datasets \cite{bochenek2025}. Taking a step further, others have examined the use of NLP techniques applied to the speeches. However, these studies are limited in scope either in terms of the techniques applied (e.g., word embeddings \cite{rodriguez2022word}) or the topic of interest (e.g., immigration \cite{card2022}).

The textual approach is especially pertinent in the Brazilian context. The political science literature has documented a notable strategic divergence between ``saying'' (speeches in the plenary, aimed at the electorate) and ``voting'' (actions often aligned with party discipline), a phenomenon known as \textit{cheap talk} or strategic signaling \cite{Moreira2020}. This discrepancy highlights that vote analysis alone provides an incomplete picture of a politician's true positioning and priorities. Our work contributes directly to this literature by analyzing the very speeches that constitute this alternative strategic dimension.

Our work is thus situated at the intersection of computational linguistics and political science, leveraging state-of-the-art NLP to address longstanding questions about legislative behavior that vote-based analyses alone cannot answer. We move beyond earlier text-as-data methods that relied on word frequencies by employing modern, high-dimensional document embeddings to capture the rich semantic nuances of parliamentary speech. Furthermore, our methodology combines a diachronic analysis of discursive evolution with a synchronic mapping of its underlying structures, offering insights that complement static, vote-based analyses. By applying this advanced semantic and temporal framework to the formal, deliberative discourse of the Brazilian Congress, our study reveals the complex, non-partisan alignment structures that remain invisible to both traditional vote-based methods and analyses of less formal political communication \cite{Baptista2021}.

\section{Methodology}

\begin{figure*}[t!]
 \centering
\includegraphics[width= 0.7\linewidth]{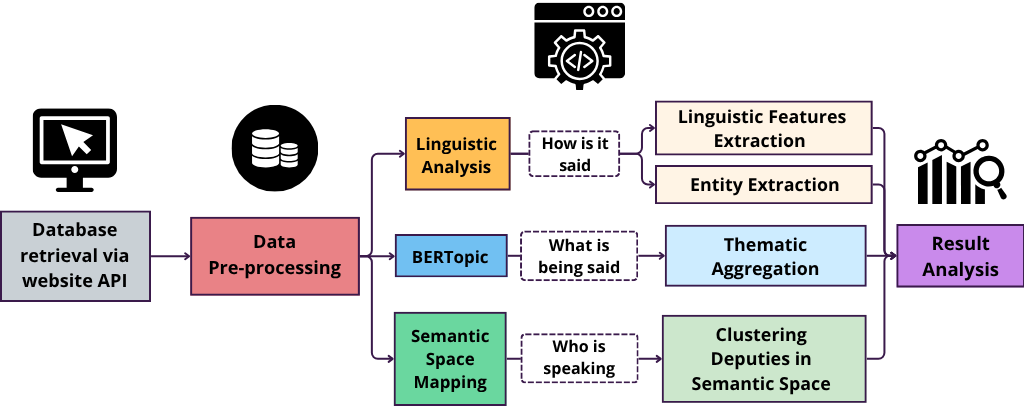}
 \caption{The multi-faceted analytical pipeline, from data retrieval to the three parallel analysis fronts: thematic, semantic, and stylometric.}
 \label{fig:pipeline}
\end{figure*}

This section presents the methodology proposed, detailing the data collection, preprocessing, modeling, and analysis steps, as illustrated in Figure~\ref{fig:pipeline}. Each step is described in detail below. While we illustrate certain tasks using examples from our case study, the approach is general and can be applied to other contexts, with minor adjustments to account for specific characteristics of the input text.

\subsection{Dataset Collection}
\label{sec:dataset}

We collected parliamentary speeches delivered in the Brazilian Chamber of Deputies between January 2003 and May 2025\footnote{The year 2025, although incomplete, was included in the analysis as it already contains 9,995 speeches, a number comparable to other years, such as 2018 (10,976) and 2024 (11,853).} from the official Open Data Portal of the Chamber\cite{camara_open_data_portal}, a public repository that provides structured access to legislative information. The collection was performed using a RESTful API provided by the Chamber of Deputies portal.

To comprehensively map the political discourse landscape, we designed a year-by-year collection strategy consisting of two main steps. First, we compiled a complete list of all deputies who held an active mandate during our analysis period. Second, for each deputy listed in the first step, we collected information from three main categories: (i) \textit{Speaker-Specific Data}, which includes detailed biographical information about each deputy who delivered at least one speech. This data comprises their unique ID, full name, gender, date of birth, municipality and state of birth, and declared level of education; (ii) \textit{Political and Temporal Context}, which refers to each speech delivered and includes: the year it occurred, the specific legislature, the party affiliation of the deputy who delivered it, and the state the deputy represented; (iii) \textit{Speech-Specific Metadata}, comprising attributes of the speech itself, including its exact start time, official title, the type of speech, the phase of the legislative session it was part of\footnote{The phases of the speech refer to the regimental moments of the plenary session. The          ``{\it Expediente}'' (which includes the ``{\it Pequeno Expedient}'' and the ``{\it Grande Expediente}'') is intended for free-topic speeches and political statements. The ``{\it Ordem do Dia}'', in turn, is the main phase dedicated to the debate and voting of legislative proposals. The other categories (Tribute, General Committee, etc.) denote sessions with specific purposes.}, a brief official summary, associated keywords, and a direct URL to the source text.


The resulting corpus gathers 453,280 speeches, delivered by 1,980 distinct deputies between 2003 and 2025. This corresponds to an average of 19,708 speeches and 497 active speakers per year. Each parliamentarian spoke on average 229 times, although the most prolific reached 8,631 interventions, and the texts contain an average of 439 words. The speeches are distributed across more than 30 parties, with  PT (Workers’ Party), MDB (Brazilian Democratic Movement), and PSDB (Brazilian Social Democracy Party) leading in volume. This link between text and structured metadata enables the central analyses of this study: thematic evolution, clustering of deputies by discursive similarity, and mapping of semantic trajectories.

\subsection{Data Preprocessing}

The entire corpus was  preprocessed for topic modeling and semantic analysis by cleaning textual artifacts, standardizing metadata, and isolating coherent monologues from dialogues or interruptions.

The first step was the isolation of single-speaker monologues. Congressional transcripts often contain interpellations or dialogues where multiple deputies speak within the same record. To ensure that each document represented the speech of a single deputy, we implemented a rule-based filter. Using regular expressions, we identified all speaker introductions in a given transcript. Specifically, our pattern was designed to match a sequence composed of: (1) a standard speaker title such as ``\textit{O Sr.}'' (Mr., in English) or ``\textit{A Deputada}'' (Congresswoman); (2) the deputy's name; (3) an optional block in parentheses, such as \textit{``(PT - PR)''}, for the party (PT) and state abreviation (PR or Paraná state); and (4) a terminal dash or hyphen. If more than one speaker was detected, the text was truncated at the exact point where the second speaker's speech began. With this approach, we exclusively preserved the first continuous segment of the main speaker's discourse. Although this methodological choice discards the remainder of the speech if the main speaker resumes later, it was fundamental to isolating a cohesive textual signal and unequivocally attributing it to a single deputy. After this, the initial speaker identification tag (e.g., ``\textit{O Sr. Deputado}'' or Congressman) was removed from the beginning of each speech.

Next, we performed extensive text normalization to remove content irrelevant to substantive political discourse. A comprehensive regular expression was used to eliminate common parliamentary formalities, honorifics, and forms of address, such as \textit{``Vossa Excelência''} (Your Excellency), \textit{``Senhor Presidente''} (Mr. President), and \textit{``nobres colegas''} (noble colleagues). Additional normalization steps included: (i) removing parenthetical asides and legislative annotations (e.g., ``nº 123''); (ii) standardizing all dashes and special characters to spaces; (iii) removing all punctuation and non-alphanumeric characters, retaining only letters and numbers; (iv) normalizing all whitespace (including line breaks and multiple spaces) to a single space; (v) converting the entire corpus to lowercase to ensure term consistency; and (vi) identifying deputies' names and standardizing them to title case.

To enable a precise longitudinal analysis, the names of political parties were corrected to reflect their current official designations, mapping historical names to their successors (e.g., PMDB (Party of the Brazilian Democratic Movement)  was renamed  MDB (Brazilian Democratic Movement) in 2018, whereas PFL (Liberal Front Party) was renamed DEM (Democrats) in 2007).  Furthermore, we established a minimum threshold of 15,000 speeches per political party. This filtering resulted in the selection of the 11 parties analyzed in this study: PT, MDB, PSDB, PL, DEM, UNIÃO\footnote{As it is the result of a  recent  merge of two parties (2021), we included UNIÃO separately to preserve the temporality of the analysis}, PP, PSB, PCdoB, PDT, and PSOL.  The choice of this threshold allows us to focus on the most prominent parties, while ensuring a sufficient sample size and maintaining a balanced distribution of parties from different points on the political spectrum.
The final corpus used in this study was limited to include exclusively the speeches from the parties of  interest, comprising a total of 354,719 pronouncements from 1582 deputies. The complete names and main ideological orientations of the selected parties along with the total number of deputies and  speeches are listed in Table~\ref{tab:party_profile} in the appendix.

\subsection{Stylometric and Linguistic Analysis}
\label{sec:linguistic_analysis}

Following the “how it is said” branch of figure \ref{fig:pipeline}, we first analyze the stylistic and structural evolution of parliamentary speeches over time. To quantify the structural properties of parliamentary discourse, we conducted a comprehensive stylometric and linguistic feature analysis. For the stylometric analysis, we used metrics of readability and lexical diversity. For the linguistic analysis, we focused on syntactic structure and named entity recognition. A brief description of these metrics is provided below:

\begin{enumerate}
\item \textbf{Readability:} We assessed the complexity of the discourse using the Flesch Reading Ease score \cite{flesch1948new}, a metric that quantifies the readability of a text based on two main variables: the average sentence length (total words divided by total sentences) and the average word complexity (total syllables divided by total words). The result is a score on a scale that typically ranges from 0 to 100. However, the formula can produce negative values for texts of extreme complexity, such as those with very long sentences and polysyllabic vocabulary, which is common in legal and parliamentary discourse. Therefore, a negative score, as observed in our results, simply indicates an exceptionally difficult-to-read text.

\item \textbf{Lexical Diversity:} The lexical richness of each speech was measured using the Type-Token Ratio (TTR) \cite{muniz2025rhetorical}, which is the ratio of the number of unique words (types) to the total number of words (tokens). Its value ranges from 0 to 1. The closer to 1, the greater the lexical diversity, indicating that the author uses a broad vocabulary and avoids repetition. The closer to 0, the lower the diversity, suggesting a more repetitive discourse with a limited vocabulary.

\item \textbf{Syntactic Structure:} To understand the grammatical composition of the discourse, we performed Part-of-Speech (POS) tagging on each token. We then calculated the relative frequency of four main grammatical categories: nouns, verbs, adjectives, and adverbs. The interpretation of these percentages reveals the discourse style: a high frequency of nouns tends to indicate a more nominal, informational, and abstract style; a high frequency of verbs suggests a more verbal, narrative, and action-oriented style; and a high frequency of adjectives and adverbs points to a more descriptive, evaluative, and subjective discourse.

\item \textbf{Named Entity Recognition (NER):} We applied NER to identify and categorize mentions of real-world entities in the text, focusing on people (PER), organizations (ORG), and locations (LOC/GPE). The analysis of this metric does not result in a single score but in frequency counts that indicate the salience of certain actors or themes.
\end{enumerate}

Each metric was calculated for each individual speech and subsequently aggregated annually. All linguistic processing, including POS tagging and NER, was performed with the spaCy library, using the pre-trained Portuguese language model pt\_core\_news\_lg \cite{honnibal2020spacy}. This diachronic approach allows for tracking the evolution of rhetorical style in the Brazilian Chamber of Deputies.

  \subsection{Thematic Analysis of Speeches}
\label{sec:topic_modeling}

We then turn to the ``what is being said'' branch of Figure~\ref{fig:pipeline}, which captures the thematic agenda of parliamentary discourse through topic modeling. To identify the latent thematic structures in the parliamentary debates, we employed BERTopic \cite{grootendorst2022bertopic}, a state-of-the-art topic modeling technique. This model has been widely adopted due to its ability to capture subtle meanings and its flexibility in automatically determining the number of topics present in the corpus, which is ideal for analyzing diverse and evolving political discourse.

To increase the coherence of the extracted topics, we first identified and joined common multi-word expressions into single tokens, aiming at enriching the vocabulary through bigrams that highlight some central themes of discussion (e.g., ``\textit{reforma tributária}'' or tax reform).  To that end, we trained a bigram detector on the tokenized documents from each year using the \textit{Phrases}\cite{gensim_phrases} model from the Gensim library.

Next, we applied the BERTopic model \cite{grootendorst2022bertopic}. The process begins by converting the various speeches from each year into vector representations using the sentence-transformer model \textit{distiluse-base-multilingual-cased-v2} \cite{reimers-2019-sentence-bert}. The choice of this model, over alternatives focused on Portuguese such as BERTimbau \cite{souza2020bertimbau}, is justified by its \textit{sentence-transformer} architecture. Such an architecture is optimized to capture the semantic context of the sentence as a whole, generating a cohesive representation for the entire text, which is more suitable for our analysis of the speeches. After vectorization, BERTopic internally applies the UMAP technique \cite{UMAP} to reduce the dimensionality of the vectors and then the HDBSCAN algorithm \cite{HDBSCAN} to group semantically similar speeches into clusters. Finally, a c-TF-IDF technique \cite{grootendorst2022bertopic} is used to extract the most representative keywords from each cluster, which constitute the topics.

The model's parameterization was determined through an iterative tuning process aimed at optimizing the quality and interpretability of the topics. We explored a range of values for key hyperparameters, focusing on `min\_topic\_size` (the minimum number of documents per topic) and the vectorizer's `min\_df` (the minimum document frequency for a word). The final selection was guided by a two-fold evaluation criterion: a qualitative assessment of the thematic coherence and distinctiveness of the resulting topics, and a quantitative monitoring of the outlier ratio (the percentage of documents not assigned to any topic). The goal was to minimize outliers without forcing documents into poorly defined, heterogeneous clusters. The chosen configuration — using a vectorizer with `min\_df=10` and setting BERTopic's `min\_topic\_size=20` — yielded the most stable and semantically meaningful thematic structure across the different years of the corpus, providing an effective balance between granularity and interpretability.

The annual modeling resulted in a large volume of granular topics, necessitating an aggregation step to obtain a high-level view. This thematic meta-analysis task, which requires the synthesis of concepts and reasoning about the socio-political domain, was conducted with the assistance of a Large Language Model (LLM), specifically Claude 3 Sonnet. To improve transparency and reproducibility, we report in Appendix~\ref{app:llm_prompt} the exact prompt used in the LLM-assisted thematic aggregation step. This prompt was used to guide the identification of six macro-themes from the yearly BERTopic outputs and the subsequent contextual lexical expansion of each category. In the first phase, the model received the complete list of annual topics with their respective keywords and was instructed to identify six broad and socio-politically meaningful macro-themes for the Brazilian context. The identified macro-themes were: Politics and Administration, Economy and Finance, Public Security and Justice, Health, Human Rights and Society, and Development and Infrastructure. 

From these macro-themes, we manually defined an initial set of seed keywords for each. In the second phase, the LLM was tasked only with performing a contextual lexical expansion of these manually defined seed sets. The model correlated the provided granular topics with the semantic patterns internalized during training to infer additional terms that, in the Brazilian socio-political domain, are likely to co-occur with the seed keywords. This process enriched the dictionary, establishing a broader and more contextually aligned thematic vocabulary, which provided the basis for the subsequent keyword-overlap classification. Finally, we used this final dictionary to automatically classify each granular topic into one of the six macro-themes, based on keyword overlap. The result of this approach is a series of annual thematic ``snapshots," revealing how the focus of parliamentary debate changes over time. 

To ensure the validity and robustness of this classification framework, we performed a two-step manual validation. First, we conducted an internal validation by reviewing a random sample of the granular topics assigned to each of the six macro-themes. This process indicated that the high-level categories provided a consistent and interpretable aggregation of the underlying topics generated by BERTopic. Second, we performed an external validation by cross-referencing the annual prominence of these macro-themes with the historical record of major socio-political events in Brazil. This confirmed that the thematic fluctuations captured by our model aligned with well-documented national crises and policy debates, thus grounding our computational findings in the real-world context.

\subsection{Analysis of the Semantic Space of Speeches}
\label{sec:embeddings}

Finally, we address the ``who speaks in discursively similar ways'' branch of Figure~\ref{fig:pipeline} through the clustering of deputies in semantic space. To compare the semantics of the speeches delivered by the deputies and to identify thematic or rhetorical convergences and divergences, we constructed a high-dimensional semantic space $\mathcal{S}^n$. In short, each individual speech $s_j$ from a deputy $d_i$ in our dataset is embedded into this space through its vector representation $x_j$. To provide a \textit{semantic fingerprint} for each deputy, an average embedding vector $x_i$ is defined as the mean of all speech embeddings associated with that deputy.

Then, to mitigate the {\it curse of dimensionality }\cite{CurseOfDimensionality} and optimize group detection, the set of average vectors was projected into a lower-dimensional space before applying a clustering algorithm, which was used to discover groups of aligned speeches and to allow for the characterization of their profiles with intra-cluster metrics. It should be noted that this analysis is synchronic and focuses on the consolidated discursive profile of each deputy from 2003 to 2025, which is distinct from the diachronic thematic analysis.

\subsubsection{Defining the Semantic Space}
To encode the semantic space, we applied the embedding model \textit{Linq-AI-Research/\\Linq-Embed-Mistral} \cite{LinqAIResearch2024}, a state-of-the-art model built on a 7-billion-parameter Mistral architecture. Its ability to handle large context windows (up to 32,768 tokens) was a key factor in our choice, as our goal was to generate representations for entire speeches without resorting to truncation or segmentation, thus preserving valuable semantic context. We used the Sentence-Transformers\cite{sentence_transformers_website} library in Python to run the model and generate a total of 354,719 vectors of 4096 dimensions.
Next, for each deputy, we calculated their \textit{semantic fingerprint}, denoted as $x_i$.

\subsubsection{Clustering  Semantic Fingerprints}
To define clusters of deputies with similar arguments and linguistic styles, we adopted the same techniques applied to the thematic analysis of speenches, namely UMAP for dimensionality reduction and  HDBSCAN  for cluster extraction.

To set the hyperparameters, we adopted an automated approach using the Optuna framework \cite{optuna_2019}. We defined an objective function aimed at maximizing the Silhouette Score \cite{Silhouette_Score}, a widely used cluster quality metric that captures both cohesion (how close deputies are to others in the same cluster) and separation (how distant they are from deputies in other clusters). To avoid trivial solutions, such as a simplistic left-right split that would obscure the core political complexity of this research, we incorporated a constraint into the objective function: hyperparameter configurations that resulted in an excessively low number of clusters were rigorously penalized during optimization. The optimization process explored a wide range of parameters for UMAP and HDBSCAN over 1,024 trials. Specifically, we tuned UMAP's \textit{n\_neighbors} (5-100), \textit{n\_components} (5-100), and \textit{min\_dist} (0.0-0.5), alongside HDBSCAN's \textit{min\_cluster\_size} (5-30), \textit{min\_samples} (1-10), and \textit{cluster\_selection\_epsilon} (0.0-0.5). This comprehensive search led to the following optimal parameterization: the number of neighbors (\textit{n\_neighbors}) and components (\textit{n\_components}) for UMAP were set to 5 and 10, respectively. For HDBSCAN, the parameters \textit{min\_cluster\_size}, \textit{min\_samples}, and \textit{cluster\_selection\_epsilon} were set to 6, 10, and 0.15, respectively.

\subsubsection{Profiling the Semantic Fingerprint Clusters}
We profiled each cluster based on three attributes of the deputies: (i) their party affiliation\footnote{The most recent affiliation of each deputy was considered.}; (ii) the state they represent; and (iii) their gender. Party affiliation helps reveal whether a given semantic cluster constitutes a party monolith or reflects an inter-party alignment. State representation reveals possible geographical influences, indicating cases where deputies align their speeches with a regional agenda. Gender can also help explain the formation of specific clusters, especially concerning sensitive topics like abortion.

To characterize possible partisan and geographical influences on the deputies' speeches, we employed the Herfindahl-Hirschman Index (HHI), a standard measure of concentration \cite{rhoades1993herfindahl}. It was adapted to calculate party concentration within each cluster (henceforth, partisan HHI or pHHI), using the formula $pHHI_c = \sum_{i=1}^{n} p_{i,c}^2$, where for a given cluster $c$, $n$ is the number of unique political parties present, and $p_{i,c}$ is the fraction of deputies in that cluster belonging to party $i$. By squaring each proportion, the index gives disproportionately greater weight to parties with a larger presence, making it highly sensitive to hegemonic actors. Values close to 1.0 indicate a monopoly by a single political party, while values close to 0 signify high partisan fragmentation. Based on party affiliation, we calculated an average ideological score for each cluster. For this, we used the categorical classification of parties (into left, center, and right) proposed by \citet{zucco2024}. Then, to allow for quantitative analysis, we converted these categories into a numerical scale, assigning the value -1 to left-wing parties, 0 to centrist parties, and 1 to right-wing parties. Analogous to the pHHI, a regional HHI (rHHI) was applied to the five macro-regions of Brazil to determine whether a cluster had a national scope or a strong regional concentration. A high rHHI indicates that the discourse of a cluster is predominantly associated with a single region. Finally, the gender composition of each cluster was analyzed by calculating the percentage of female deputies relative to its total members.

The experiments were conducted on a single workstation equipped with an AMD Ryzen 9 7950X 16-Core CPU, 128GB of RAM, and a single NVIDIA RTX 6000 Ada Generation GPU. The most computationally intensive step was the generation of document embeddings for over 350,000 speeches. The entire analysis pipeline, took approximately 90 hours to complete.

\section{Analysis of the Speeches}


\subsection{Style and Structure Over Time}
\label{sec:linguistic_results}

\begin{figure*}[t!]
    \centering
    \includegraphics[width=0.7\linewidth]{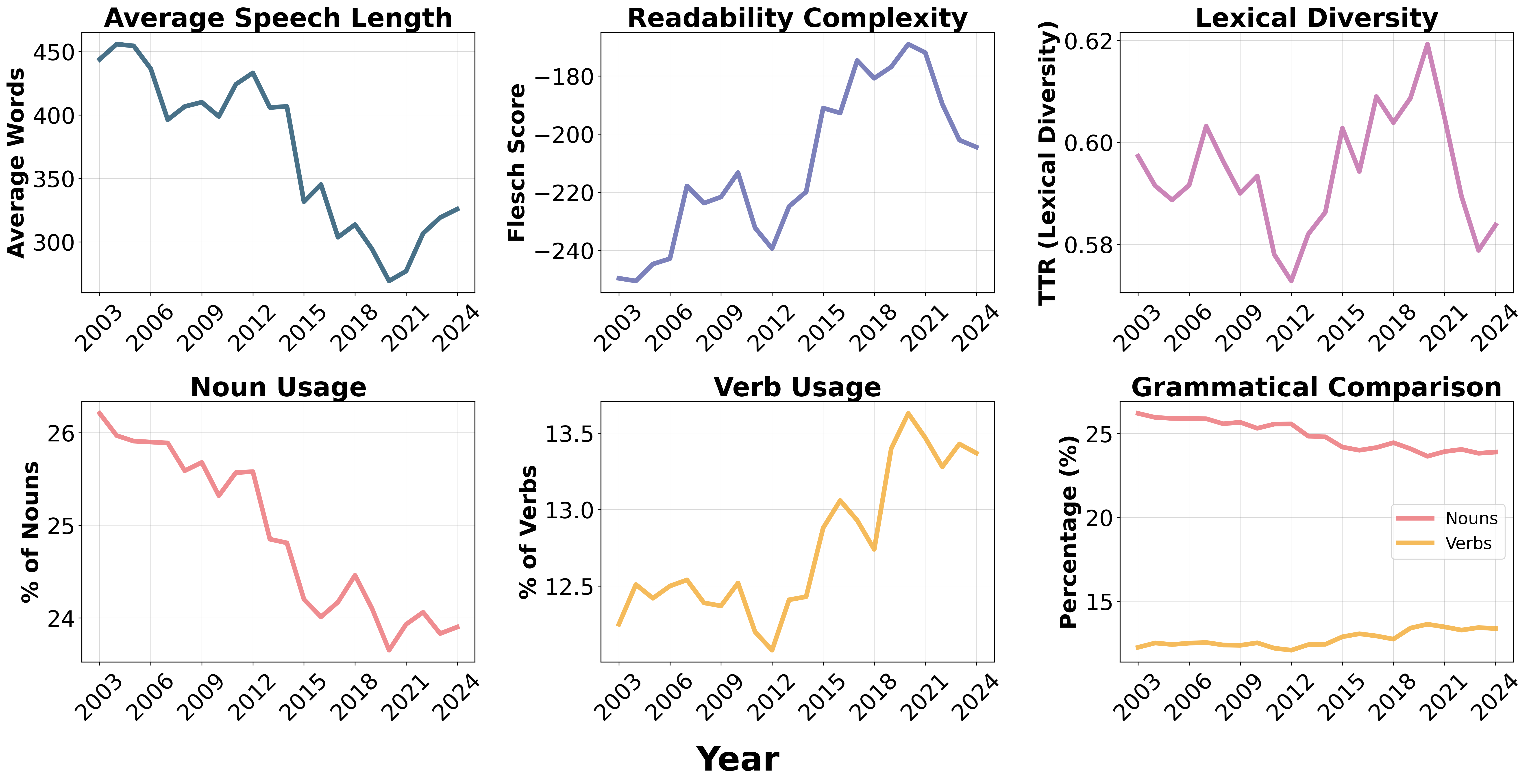} 
    \caption{Evolution of stylometric (top row) and grammatical (bottom row) metrics of parliamentary discourse (2003-2024).}
    \label{fig:analise_linguistica_completa}
\end{figure*}

The analysis of the linguistic and stylistic characteristics of the speeches reveals a significant evolution in the style of parliamentary communication. For this specific time-series analysis, we focused on the period from 2003 to 2024. Unlike our thematic and semantic analyses, this approach relies on stable annual averages. Including the partial data from 2025 could therefore introduce bias and skew the longitudinal trends. Figure~\ref{fig:analise_linguistica_completa} presents the trajectory of six key metrics over this period of complete years.

The top row of Figure~\ref{fig:analise_linguistica_completa} details the stylometric metrics. The most visible trend is a movement towards conciseness. The average length of speeches (left graph) peaked at 456 words in 2004 and has since followed a downward trajectory, reaching its lowest point of 269 words in 2020. This change is mirrored by readability (center graph): the Flesch Index, which indicates exceptionally difficult-to-read text throughout the period, reached its lowest (most complex) point of -250.48 in 2004 and its peak (least complex) of -169.07 in 2020, coinciding with the year of the shortest speeches. Interestingly, this structural simplification did not imply an impoverishment of the vocabulary. The lexical diversity (TTR), in the right graph, remained remarkably stable and high, indicating that, although shorter, the speeches maintained a rich vocabulary.

The bottom row of the figure focuses on grammatical structure, revealing a subtle but consistent shift in argumentation style. A gradual decline in the proportion of nouns (left graph) is observed, with a corresponding increase in verbs (center graph). The direct comparison in the last graph makes this trend reversal clear, which becomes more pronounced after 2015. This transformation from a more "nominal" to a more "verbal" style suggests a shift towards more direct, action- and process-focused communication, at the expense of more abstract and conceptual language, which is consistent with theories of the mediatization of politics \cite{strömbäck2008four}.


While stylistic and grammatical metrics capture how deputies speak, Named Entity Recognition (NER) reveals who and what they speak about, serving as a thermometer of political attention. Table~\ref{tab:entity_evolution} summarizes the salience of key figures across periods.

The analysis confirms the Executive Branch as the main ``discursive center of gravity'', with the incumbent president (Lula, Dilma Rousseff, Jair Bolsonaro) consistently being the most mentioned figure, whether as a point of support or opposition. Furthermore, the table serves as a historical record of crises and institutional power. 
 The prominence of Eduardo Cunha (Speaker of the Chamber of Deputies, 2015–2016) and Arthur Lira (Speaker of the Chamber, 2021–2023) highlights periods when the office of the Chamber’s presidency wielded crucial influence.
Similarly,  the growing prominence  from 2022 onwards of  Alexandre de Moraes, Chief Justice of Brazil’s Supreme Federal Court,  reflects the intensification of the judicialization of politics.

Finally, the data captures how social agendas reverberate in the plenary. The prominence of councilwoman and human rights activist {\it Marielle Franco}, whose assassination in 2018 became a symbol of the struggle against political violence in Brazil\cite{guardian_marielle_franco}, illustrates how tragic events enter legislative debate. Likewise, the repeated mentions of {\it Paulo Gustavo}, a popular comedian whose name was given to a cultural emergency law after his death from COVID-19, and Aldir Blanc, a celebrated songwriter honored through another cultural funding law, highlight the role of artists in shaping political discussions. References to Pope Francis as a moral authority on human rights further underscore the plenary as an arena where society’s broader agendas are both debated and contested.


\begin{table*}[t!]
\centering
{\small
\setlength{\tabcolsep}{3pt}
\begin{tabular}{@{} p{4.5cm} c p{10cm} @{}}
\toprule
\textbf{Figure (Category)} & \textbf{Peak Period} & \textbf{Associated Context} \\
\midrule
\multicolumn{3}{@{}l}{\textit{\textbf{Executive Power}}} \\
\textbf{Lula} \textit{(President)} & 2003-10, 2023-25 & Central figure of his administration; a focal point for support and opposition. \\
\textbf{Dilma Rousseff} \textit{(President)} & 2014-16 & Salience peaks during her presidency and the 2015-16 impeachment crisis. \\
\textbf{Jair Bolsonaro} \textit{(President)} & 2019-22, 2025 & Focus of his administration and lasting post-presidency polarization. \\
\midrule
\multicolumn{3}{@{}l}{\textit{\textbf{Legislative and Judiciary Powers}}} \\
\textbf{Eduardo Cunha} \textit{(Speaker of the Chamber of Deputies )} & 2015-16 & Protagonist of the 2015-16 presidential impeachment process. \\
\textbf{Arthur Lira} \textit{(Speaker of the Chamber of Deputies)} & 2021-23 & Key figure to governability and ``secret budget'' controversies. \\
\textbf{Alexandre de Moraes} \textit{(Chief Justice of Supreme Federal Court)} & 2022-25 & Key figure during conflicts over elections, anti-democratic acts, and the judicialization of politics. \\
\midrule
\multicolumn{3}{@{}l}{\textit{\textbf{Social and Cultural Symbols}}} \\
\textbf{Marielle Franco} \textit{(Human-Rights Activist)} & 2018-19, 2024 & Icon in debates on political violence and justice after her 2018 assassination. \\
\textbf{Paulo Gustavo} and \textbf{Aldir Blanc} \textit{(Artists)} & 2022, 2025 & Names of major cultural funding laws passed after their deaths. \\
\textbf{Pope Francis} \textit{(Moral Authority)} & 2013-15, 2025 & Invoked as a moral authority on inequality, the environment, and human rights. \\
\bottomrule
\end{tabular}
}
\caption{Main Named Entities in the Speeches.}
\label{tab:entity_evolution}
\end{table*}

In summary, the linguistic and stylometric analysis of the speeches reveals a dual evolution in Brazilian parliamentary discourse over the last two decades. On the one hand, we observe a clear structural simplification: the speeches have become shorter, more direct, and have adopted a more verbal style, adapting to a faster-paced media ecosystem. On the other hand, the content analysis shows a dynamic thematic focus, where the salience of actors and institutions functions as a direct reflection of the political crises and social agendas that shaped the national agenda during the period.

\subsection{Thematic Landscape of Parliamentary Discourse}

\begin{figure*}[t!]
    \centering
    \includegraphics[width=0.7\linewidth]{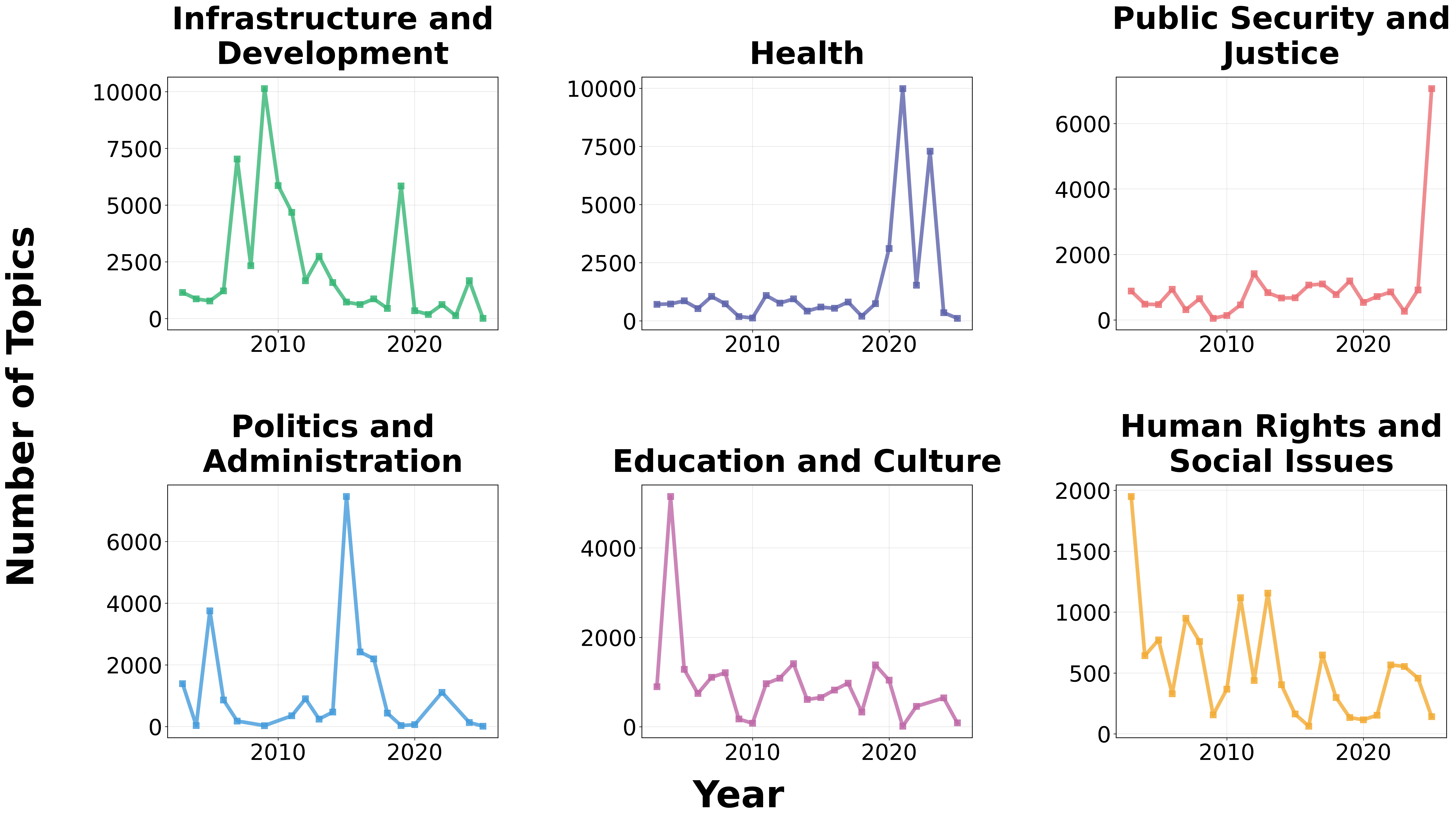} 
    \caption{Temporal evolution of the discourse for 6 macro-themes (2003-2025).}
    \label{fig:theme_evolution}
\end{figure*}
\begin{figure*}[t!]
    \centering
    \includegraphics[width=0.95\linewidth]{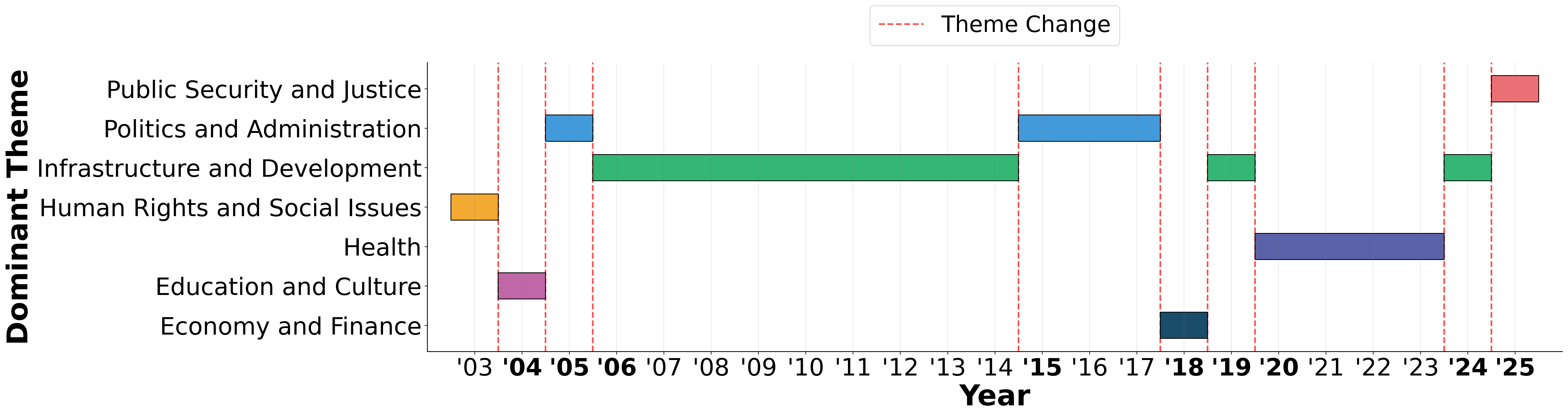}
    \caption{A timeline visualizing the dominant theme in parliamentary discourse (2003-2025).}
    \label{fig:dominant_timeline}
\end{figure*}

Our analysis of the thematic structure of parliamentary discourse, derived from topic modeling of the entire speech corpus, reveals a dynamic and highly responsive legislative agenda. We identified six distinct macro-themes that consistently organize the debates, which are presented in Table~\ref{tab:theme_keywords} along with their most representative keywords. These themes range from perennial governance topics, such as \textit{Politics and Administration}, to more event-driven issues, like \textit{Health} and \textit{Human Rights}.

\begin{table}[h!]
\centering
{\small
\setlength{\tabcolsep}{2pt}
\begin{tabular}{@{}p{0.30\columnwidth}p{0.60\columnwidth}@{}}
\toprule
\textbf{Theme} & \textbf{Main Representative Keywords} \\
\midrule
Politics and Admin. & democracy, party, pt, corruption, election \\
Public Security & violence, public security, justice, crime, police \\
Health & health, disease, ministry of health, sus, patients \\
Human Rights & children, youth, rights, women, society \\
Infrastructure & water, transport, mayor, energy, highways \\
Education and Culture & culture, students, education, teachers, school \\
\bottomrule
\end{tabular}
}
\caption{Main Macro-Themes of Parliamentary Discourse}

\label{tab:theme_keywords}
\end{table}

Although, in general terms, development and public service themes form the foundation of the debate, such a static view masks the profound shifts in legislative attention over the past two decades. The true insights emerge from analyzing the evolution of these themes over time.

\subsubsection{Evolution of Thematic Attention (2003-2025)}

The evolution of thematic attention, visualized in Figure~\ref{fig:theme_evolution}, shows that the legislative agenda is in constant flux, often reorienting itself in response to national priorities. The figure, which plots the proportional share of discourse for three key themes, reveals two defining long-term trends.

First, the blue line representing \textit{Infrastructure and Development} shows a distinct historical arc. This theme maintained high levels of salience for nearly a decade, peaking between 2007 and 2011, a period concurrent with major national investment programs like the ``{\it Programa de Aceleração do Crescimento} '' (PAC) (or Growth Acceleration Program). Since then, its prominence has been in a clear, though fluctuating, decline, indicating a shift away from large-scale development as a central focus of debate.

Second, the figure highlights the recent and dramatic rise of two other themes. The green line, representing \textit{Health}, remained a stable, mid-level concern for 17 years before experiencing an unprecedented peak between 2020 and 2023. This surge, during which \textit{Health} becomes the most dominant topic, is a clear signature of the COVID-19 pandemic. More recently, \textit{Public Security and Justice} shows a sharp upward trajectory, culminating in 2025. This suggests that concerns with security and justice have forcefully entered the agenda, displacing other long-standing priorities.

\subsubsection{The Legislative Agenda in Periods of National Crisis}

The timeline of the most dominant theme each year, presented in Figure~\ref{fig:dominant_timeline}, offers an even clearer view of how national crises dictate the legislative focus. The figure reveals distinct ``eras'' in which the political agenda was captured by a specific theme.

The long {\it Infrastructure Era}, from 2006 to 2014, is clearly visible, showing a sustained period of focus on development. This period is punctuated by two moments of intense institutional crisis, where Politics and Administration takes control of the agenda. The first is in 2005, corresponding directly to the ``Mensalão'',
 one of the biggest political corruption scandal in the country. The second, a more prolonged period from 2015 to 2017, aligns perfectly with the \textit{impeachment} of President Dilma Rousseff and the political turmoil of the subsequent administration.

More recently, the timeline shows the impact of the COVID-19 pandemic, which initiated a {\it Health Era} from 2020 to 2023. The abrupt shift and subsequent pivot to Public Security and Justice in 2025 underscore the Chamber's high reactivity to exogenous shocks. This analysis reveals that the t hematic agenda is not a stable list of priorities but a dynamic battlefield of issues, where long-term trends are punctuated by sudden, crisis-driven reorientations.

\subsection{Clustering of Deputies}

\begin{figure}[t]
    \centering
    \includegraphics[width=0.7\columnwidth]{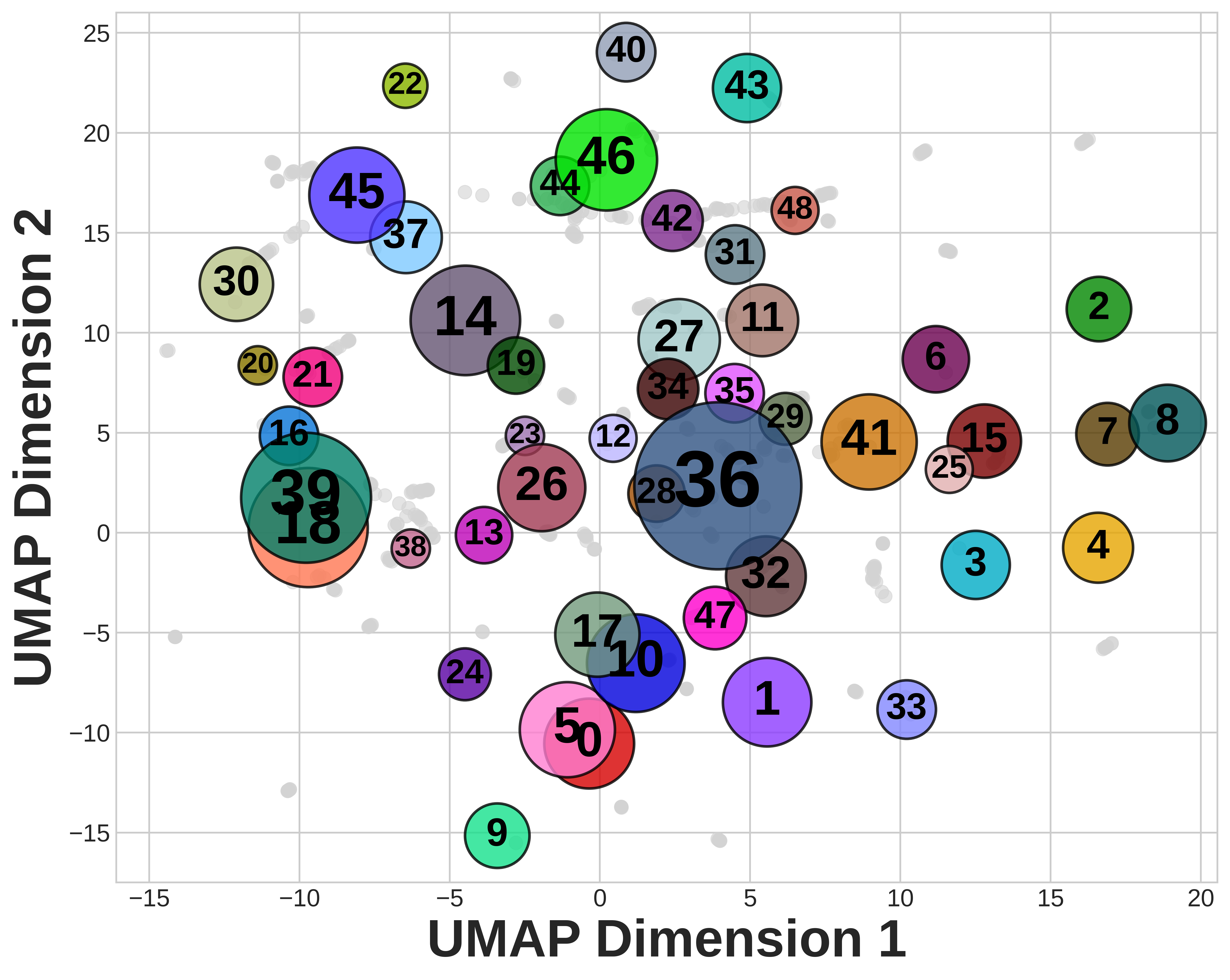}
    \caption{2D visualization of the centroids of the 49 identified discursive clusters. Each colored point represents the center of mass of a cluster, with its size proportional to the number of deputies it contains. The gray points represent deputies classified as noise.}
    \label{fig:cluster}
\end{figure}

The robustness of our semantic mapping was validated by a strong Silhouette Score of 0.647, which confirms the formation of internally cohesive and well-separated clusters of deputies. From this solid foundation, the model identified 49 distinct discursive clusters (Figure~\ref{fig:cluster})—a number that dramatically exceeds the 11 major political parties considered in this study. This high degree of granularity offers a novel, speech-based perspective that complements and expands upon previous findings of fragmentation in the Brazilian party system, which have been documented primarily through roll-call vote analyses \cite{PaperCarlos}.

While methods based on voting records are powerful for revealing the \textit{outcomes} of political negotiation, our discursive analysis provides an orthogonal view, uncovering the semantic and thematic alignments that shape representatives' public stances. This suggests that for many deputies, party affiliation is not the primary determinant of discursive alignment. Instead, the semantic space is shaped by a complex interaction of regional interests, specific policy agendas, and ideological nuances that transcend formal party structures, offering a more direct view into the forces driving political behavior, which often remain latent in the data from roll-call votes.

Furthermore, a key feature of the HDBSCAN algorithm is its ability to identify points that do not belong to any distinct \textit{cluster}, classifying them as ``noise'' or \textit{outliers}. Rather than a methodological limitation, this classification provides a valuable insight into the nature of the data itself. In total, 30.4\% of the analyzed deputies were classified as noise. There are two main explanations for this result. First, some of these \textit{outliers} correspond to deputies with a very low number of recorded speeches. For these legislators, the aggregated \textit{embedding} is derived from insufficient data, resulting in a ``semantic fingerprint'' that is not representative enough to confidently join a well-defined discursive \textit{cluster}. Second, this group also includes politicians whose discourse is generalist. By avoiding strong ideological markers or deep engagement with specific policy niches, their speeches lack the distinctive features necessary for clustering. Their \textit{embeddings}, consequently, occupy a central but indistinct region of the semantic space, failing to gravitate towards any thematic or ideological pole. The algorithm's ability to isolate these cases is a  strength because, by not forcing these deputies into ill-fitting groups, it increases the purity and interpretability of the core ideological \textit{clusters} that emerge.

An intra-cluster analysis of the 49 thematic groupings was conducted to characterize the underlying factors driving their formation. Our results indicate that, in addition to party affiliation, clustering is also impacted by geographical characteristics and gender-related factors.

Specifically, a quantitative analysis reveals that most \textit{clusters} are not simple party monoliths. The average partisan Herfindahl-Hirschman Index (pHHI) across all \textit{clusters} is only 0.28, indicating a high degree of political fragmentation. Indeed, many groups, such as Cluster 36, include representatives from almost all parties analyzed in this study and exhibit a pHHI as low as 0.13. The exceptions to this rule are found at the ideological extremes. For example, Cluster 40 (PSOL) and Cluster 16 (PCdoB) show exceptionally high pHHIs (0.87 and 0.76, respectively) and are ideologically located at the left pole (average ideological score of -1.0). This demonstrates that while strong party cohesion exists, it is rare and confined mainly to ideologically oriented groups.

Our analysis reveals that a significant number of \textit{clusters} have a strong regional character. In fact, 23 of the 49 thematic groupings show high regional concentration, defined as having a regional HHI (rHHI) score above 0.7. Ten of these clusters reach the maximum rHHI value of 1.0, indicating complete regional dominance. This phenomenon extends across all of Brazil; for example, Cluster 13 is composed exclusively of deputies from the South, Cluster 28 from the Northeast, and Cluster 4 from the North, each indicating a potent and regionally defined discursive identity. This pattern strongly suggests the formation of regionally aligned discursive groups, where local interests and political priorities create a common discourse that transcends party affiliations.

Finally, our analysis also reveals several smaller \textit{clusters} with an overwhelmingly high representation of women. For instance, Clusters 37 and 19 are over 90\% female—a stark contrast to the Chamber's average female membership of approximately 15\% during the analyzed period. More surprisingly, these female-dominated \textit{clusters} are politically diverse, with representatives from up to seven different parties. This finding indicates that gender can be a powerful factor in creating discursive affinity, independent of the ideological spectrum. This pattern suggests the existence of a distinct female political discourse or a tendency for women leaders to form cohesive, cross-party semantic groups, a phenomenon that merits further qualitative investigation.

\section{Discussion}

Our analysis of over two decades of parliamentary speeches, enabled by a novel computational framework, reveals a complex political landscape where alignments are often driven by a nuanced interplay of ideology, regional identity, and gender that supersedes formal party structures. This work's main contribution is to provide a multidimensional account of parliamentary discourse through a scalable computational framework: we introduce and validate a scalable approach that triangulates stylometric, thematic, and semantic analyses of institutional corpora. Each component captures a distinct dimension of parliamentary discourse: stylometric analysis reveals how speeches are constructed, thematic analysis reveals what issues organize parliamentary debate, and semantic analysis reveals who speaks in discursively similar ways. Taken together, these dimensions allow us to study parliamentary discourse as a phenomenon of form, content, and alignment, rather than reducing it to any single textual property. By applying this framework to the Brazilian Congress, we demonstrate its power to decode the rhetoric, strategic signaling, and identity-based coalitions that remain invisible to traditional roll-call vote or social media analyses, opening new avenues for comparative legislative studies worldwide.

We first observe a significant stylistic evolution in parliamentary communication. The clear trend toward shorter, more readable, and verbally-oriented speeches corroborates theories on the mediatization of politics, suggesting an adaptation to a faster-paced media ecosystem where action-oriented language gains more traction \cite{muniz2025rhetorical}. Crucially, however, this structural simplification does not equate to a loss of substantive depth. Our thematic analysis demonstrates that the legislative agenda remains highly responsive to national crises, from corruption scandals to the COVID-19 pandemic, reflecting an institution whose focus is dynamically shaped by exogenous shocks, consistent with models of institutional agenda-setting \cite{ferreira2020agenda, ouverney2022legislative}.

The semantic analysis provides the most granular insights into these discursive alignments. The identification of 49 distinct semantic clusters, a number far exceeding that of the political parties studied, confirms that party affiliation is an incomplete predictor of political speech. Many of these clusters exhibit strong geographic cohesion, suggesting a form of "discursive federalism" where regional agendas forge cross-party alliances. This finding empirically grounds literature that points to overlapping axes of representation in Brazilian politics, demonstrating that for many legislators, local interests are a powerful driver of discursive identity \cite{delaporte2023female_representation, boas2019congruence}.

Furthermore, our model uncovers a compelling gender dimension in legislative discourse. We identified several clusters with an overwhelmingly high proportion of female deputies that are, surprisingly, politically heterogeneous, spanning multiple parties from across the ideological spectrum. This suggests that gender can operate as a primary organizing principle for discursive affinity, potentially giving rise to a distinct "female political discourse." This quantitative finding resonates with feminist political literature, which has long argued that women in politics often prioritize and frame issues differently, particularly those concerning social rights, health, and education \cite{miguel2009gender_discourse, brigagao2024gender_violence_parliament}, and our work provides a new methodology to study this phenomenon at scale.

In conclusion, by articulating three complementary dimensions—form, content, and alignment—our approach builds a rich, multidimensional representation of parliamentary behavior. It contributes a robust methodology for transforming raw institutional data into actionable political insight, offering an internal lens on deliberative processes that complements and enriches findings from voting records or public-facing social media data.



\section{Limitations and Future Work}

Our study, however, has limitations, both methodological and interpretive. Methodologically, the analysis does not link speeches to their specific legislative context (e.g., the bill being debated), nor does it capture the interactive dynamics of parliamentary debate. Interpretively, we caution against the potential misuse of our findings. The identification of group-based discursive patterns, particularly around regional or gender identity, could be instrumentalized to create or reinforce stereotypes. We therefore emphasize that our results should be seen as a map of complex group dynamics, not as essentialist labels for individuals or their politics.

These limitations highlight promising avenues for future research. A key next step is to link discursive content to legislative action, connecting speeches to specific bills to model the impact of rhetoric on policy outcomes. Another significant computational challenge lies in modeling deliberative dialogue, moving beyond monologues to analyze the persuasive exchanges and interruptions that define debate. Furthermore, evaluating our methodology using alternative language models, particularly those specifically pre-trained on Brazilian Portuguese, constitutes an important step to assess the robustness of our findings. Moreover, although our analysis focuses on speeches from the Brazilian Chamber of Deputies, the proposed approach is general and can be applied to other contexts, for instance, by leveraging recently published curated datasets from the U.S. Congressional Record \cite{bochenek2025}. Finally, our work lays the groundwork for developing interactive platforms that would enable journalists, researchers, and the public to explore these discursive patterns, fostering greater transparency and civic engagement.

\section{Conclusions}

In this paper, we propose a novel NLP-based methodology for the large-scale, longitudinal analysis of legislative discourse, applying it to over two decades of speeches from the Brazilian Chamber of Deputies. Our study uncovers a legislative agenda that is highly reactive to national crises and reveals a complex map of semantic alignments where regional and gender identities often prove more salient than partisan affiliation. By doing so, it validates our approach as an effective method for extracting nuanced political dynamics from vast, unstructured institutional text.

Ultimately, this research demonstrates how computational social science can illuminate the inner workings of democratic institutions. By transforming vast textual archives into interpretable maps of thematic priorities and political coalitions, we offer a methodology that not only deepens academic understanding of representation but also provides new tools for public accountability in the digital age.

\section*{Acknowledgements}
This work was supported by the Brazilian funding agencies CNPq (Conselho Nacional de Desenvolvimento Científico e Tecnológico), CAPES (Coordenação de Aperfeiçoamento de Pessoal de Nível Superior) and FAPEMIG (Fundação de Amparo à Pesquisa do Estado de Minas Gerais), as well as by the National Institute of Science and Technology in Responsible Artificial Intelligence for Computational Linguistics and for Information Treatment and Dissemination (INCT-TILD-IAR).

\bibliography{references}
\section{Paper Checklist}

\begin{enumerate}

\item For most authors...
\begin{enumerate}
    \item  Would answering this research question advance science without violating social contracts, such as violating privacy norms, perpetuating unfair profiling, exacerbating the socio-economic divide, or implying disrespect to societies or cultures?
    \answerYes{Yes}
  \item Do your main claims in the abstract and introduction accurately reflect the paper's contributions and scope?
    \answerYes{Yes}
   \item Do you clarify how the proposed methodological approach is appropriate for the claims made? 
    \answerYes{Yes}
   \item Do you clarify what are possible artifacts in the data used, given population-specific distributions?
    \answerYes{Yes}
  \item Did you describe the limitations of your work?
    \answerYes{Yes}
  \item Did you discuss any potential negative societal impacts of your work?
    \answerYes{Yes}
      \item Did you discuss any potential misuse of your work?
    \answerYes{Yes}
    \item Did you describe steps taken to prevent or mitigate potential negative outcomes of the research, such as data and model documentation, data anonymization, responsible release, access control, and the reproducibility of findings?
    \answerYes{Yes. The paper details its primary mitigation step providing explicit interpretive guidance to prevent the misuse of findings for stereotyping. This constitutes the core of our responsible release strategy for the findings themselves. Regarding the other points: (i) data anonymization is not applicable, as the study analyzes the public records of public figures acting in their official capacity; and (ii) to ensure reproducibility, the authors commit to releasing the well-documented code upon publication, which will serve as the data and model documentation.}
  \item Have you read the ethics review guidelines and ensured that your paper conforms to them?
    \answerYes{Yes}
\end{enumerate}

\item Additionally, if your study involves hypotheses testing...
\begin{enumerate}
  \item Did you clearly state the assumptions underlying all theoretical results?
    \answerYes{Yes}
  \item Have you provided justifications for all theoretical results?
    \answerYes{Yes}
  \item Did you discuss competing hypotheses or theories that might challenge or complement your theoretical results?
    \answerYes{Yes}
  \item Have you considered alternative mechanisms or explanations that might account for the same outcomes observed in your study?
    \answerNA{NA}
  \item Did you address potential biases or limitations in your theoretical framework?
    \answerNA{NA}
  \item Have you related your theoretical results to the existing literature in social science?
    \answerYes{Yes}
  \item Did you discuss the implications of your theoretical results for policy, practice, or further research in the social science domain?
    \answerYes{Yes}
\end{enumerate}

\item Additionally, if you are including theoretical proofs...
\begin{enumerate}
  \item Did you state the full set of assumptions of all theoretical results?
    \answerNA{NA}
	\item Did you include complete proofs of all theoretical results?
    \answerNA{NA}
\end{enumerate}

\item Additionally, if you ran machine learning experiments...
\begin{enumerate}
  \item Did you include the code, data, and instructions needed to reproduce the main experimental results (either in the supplemental material or as a URL)?
    \answerYes{Yes}
  \item Did you specify all the training details (e.g., data splits, hyperparameters, how they were chosen)?
    \answerYes{Yes}
     \item Did you report error bars (e.g., with respect to the random seed after running experiments multiple times)?
    \answerNo{No. As our study is an exploratory analysis using unsupervised clustering, assessing variance across different random seeds is not central to our claims. The robustness of our findings is instead supported by the strong Silhouette Score of 0.647, which validates the quality and stability of the discovered clusters.} 
	\item Did you include the total amount of compute and the type of resources used (e.g., type of GPUs, internal cluster, or cloud provider)?
    \answerYes{Yes}
     \item Do you justify how the proposed evaluation is sufficient and appropriate to the claims made? 
    \answerYes{Yes}
     \item Do you discuss what is ``the cost`` of misclassification and fault (in)tolerance?
    \answerYes{Yes}
  
\end{enumerate}

\item Additionally, if you are using existing assets (e.g., code, data, models) or curating/releasing new assets, \textbf{without compromising anonymity}...
\begin{enumerate}
  \item If your work uses existing assets, did you cite the creators?
    \answerYes{Yes}
  \item Did you mention the license of the assets?
    \answerNo{No. The assets used are public records}
  \item Did you include any new assets in the supplemental material or as a URL?
    \answerYes{Yes}
  \item Did you discuss whether and how consent was obtained from people whose data you're using/curating?
    \answerNA{NA. The dataset consists of official transcripts of speeches delivered by elected representatives during public parliamentary sessions. As this is public record data involving public figures acting in their official capacity, individual consent for research use is not required.}
  \item Did you discuss whether the data you are using/curating contains personally identifiable information or offensive content?
    \answerYes{Yes. By design, the data contains personally identifiable information —namely the names and party affiliations of public figures (the deputies)—as this information is essential for the analysis. Furthermore, as a record of political debate, the corpus may contain language that could be considered offensive or prejudiced, which is treated as an intrinsic and organic component of the political discourse under investigation.}
\item If you are curating or releasing new datasets, did you discuss how you intend to make your datasets FAIR?
\answerNA{NA}
\item If you are curating or releasing new datasets, did you create a Datasheet for the Dataset? 
\answerNA{NA}
\end{enumerate}

\item Additionally, if you used crowdsourcing or conducted research with human subjects, \textbf{without compromising anonymity}...
\begin{enumerate}
  \item Did you include the full text of instructions given to participants and screenshots?
    \answerNA{NA}
  \item Did you describe any potential participant risks, with mentions of Institutional Review Board (IRB) approvals?
    \answerNA{NA}
  \item Did you include the estimated hourly wage paid to participants and the total amount spent on participant compensation?
    \answerNA{NA}
   \item Did you discuss how data is stored, shared, and deidentified?
   \answerNA{NA}
   
\end{enumerate}

\end{enumerate}

\onecolumn
\appendix

\section*{Appendix}

\section{Prompts Used for the LLM-Assisted Thematic Aggregation}
\label{app:llm_prompt}

To support transparency and reproducibility, we report below the prompt templates used in the two-stage LLM-assisted thematic aggregation procedure described in the subsection \textit{Thematic Analysis of Speeches}. In the first stage, the LLM was used to propose six high-level macro-themes based on the yearly BERTopic outputs. In the second stage, after we manually defined an initial set of seed keywords for each macro-theme, the LLM was used only for contextual lexical expansion. The prompts below are reported to document the role of the LLM in each step of the pipeline.

\captionsetup{type=listing}
\captionof{listing}{Prompt used in Stage 1 for macro-theme identification.}
\label{lst:llm_prompt_stage1}
\begin{lstlisting}
You are assisting with the thematic analysis of parliamentary speeches from the Brazilian Chamber of Deputies.

You will receive a list of granular topics extracted with BERTopic. Each topic contains representative keywords.

Task:
Read the full list of topics and propose exactly six macro-themes that summarize the main areas of parliamentary debate in Brazil.

Instructions:
- The categories must be broad, distinct, and socio-politically meaningful in the Brazilian context.
- The categories should be general enough to cover multiple yearly topics.
- Avoid overlapping or redundant categories.
- Avoid labels that are too generic.
- Base your reasoning only on the topic keywords provided and on broad socio-political knowledge.

For each macro-theme, provide:
- a label;
- a one-sentence description.

Input format:
Topic_ID: keyword1, keyword2, keyword3, ...

Expected output:
1. A list of exactly six macro-themes.
2. For each macro-theme:
   - label
   - short description
\end{lstlisting}

\captionsetup{type=listing}
\captionof{listing}{Prompt used in Stage 2 for contextual lexical expansion.}
\label{lst:llm_prompt_stage2}
\begin{lstlisting}
You are assisting with the thematic analysis of parliamentary speeches from the Brazilian Chamber of Deputies.

You will receive:
1. A list of macro-themes previously defined for the analysis;
2. An initial manually defined set of seed keywords for each macro-theme;
3. A list of granular BERTopic topics with representative keywords.

Task:
For each macro-theme, expand the initial seed keyword list with additional terms that are contextually related and likely to co-occur with the seed keywords in Brazilian parliamentary discourse.

Instructions:
- Preserve the meaning and scope of each macro-theme.
- Suggest only terms that are semantically related to the provided seed keywords.
- Use the granular topics as contextual support for the expansion.
- Focus on vocabulary that is plausible in the Brazilian socio-political and legislative domain.
- Do not create new macro-themes.
- Do not reclassify topics.
- Do not include overly generic or weakly related words.

Input format:
Macro-theme: [label]
Seed keywords: keyword1, keyword2, keyword3, ...
Related BERTopic topics:
- Topic_ID: keyword1, keyword2, keyword3, ...
- Topic_ID: keyword1, keyword2, keyword3, ...

Expected output:
For each macro-theme:
- Macro-theme label
- Expanded keyword list
\end{lstlisting}

\section{Supplementary Party Profile}
\label{app:party_profile}

Table~\ref{tab:party_profile} provides complementary descriptive information about the 11 political parties included in our final corpus. In addition to reporting the number of deputies and speeches associated with each party, the table includes the ideological orientation adopted throughout the paper, based on the classification by Zucco and Power (2024). This information supports the interpretation of the semantic clusters discussed in the subsection \textit{Analysis of the Semantic Space of Speeches}, particularly the analyses of partisan concentration and ideological positioning.

\begin{table*}[h]
\centering

\begin{tabular}{@{} llrrl @{}}
\toprule
\textbf{Acronym} & \textbf{Full Name} & \textbf{Deputies} & \textbf{Speeches} & \textbf{Ideology} \\
\midrule
PT      & Partido dos Trabalhadores        & 262 & 97,560 & Left \\
MDB     & Movimento Democrático Brasileiro & 312 & 45,408 & Center \\
PSDB    & Partido da Social Democracia Brasileira & 234 & 37,539 & Center \\
PL      & Partido Liberal                  & 254 & 36,346 & Right \\
DEM     & Democratas                       & 205 & 32,645 & Right \\
UNIÃO   & União Brasil                     & 67 & 1,707 & Right \\
PP      & Progressistas                    & 189 & 25,254 & Right \\
PSB     & Partido Socialista Brasileiro    & 146 & 23,361 & Left \\
PCdoB   & Partido Comunista do Brasil      & 41 & 19,075 & Left \\
PDT     & Partido Democrático Trabalhista  & 115 & 18,170 & Left \\
PSOL    & Partido Socialismo e Liberdade   & 26 & 17,654 & Left \\
\bottomrule

\end{tabular}
\caption{Descriptive profile of the 11 political parties included in the final corpus. For each party, we report its acronym, full name, number of deputies, number of speeches, and ideological orientation based on Zucco and Power (2024).}
\label{tab:party_profile}
\end{table*}

\twocolumn

\end{document}